\documentclass[11pt,a4paper]{article}
\usepackage[hyperref]{acl2020}
\usepackage{graphicx}
\usepackage{times}
\usepackage{latexsym}
\usepackage{subfigure}
\usepackage{url}
\usepackage{epstopdf}
\usepackage{amsmath}
\usepackage{multirow}
\usepackage{amssymb}
\usepackage{array}
\usepackage{booktabs}
\usepackage{algorithm}
\usepackage{algorithmic}
\usepackage{bm}
\usepackage{makecell}

\usepackage{microtype}

\aclfinalcopy 

\title{PONE: A Novel Automatic Evaluation Metric for Open-Domain Generative Dialogue Systems}

\author{Tian Lan, Xian-Ling Mao, Wei Wei, Xiaoyan Gao, Heyan Huang \\ 
   Beijing Institute of Technology \\
   Huazhong University of Science and Technology \\
\texttt{lantiangmftby@gmail.com, \{maoxl,xygao,hhy63\}@bit.edu.cn} \\ 
\texttt{weiw@hust.edu.cn} \\}
\date{}

\begin{document}
\maketitle
\begin{abstract}
    Open-domain generative dialogue systems have attracted considerable attention over the past few years.
    Currently, how to automatically evaluate them, is still a big challenge problem. 
    As far as we know, there are three kinds of automatic methods to evaluate the open-domain generative dialogue systems: 
    (1) Word-overlap-based metrics; (2) Embedding-based metrics; (3) Learning-based metrics. 
    Due to the lack of systematic comparison, it is not clear which kind of metrics are more effective.  
    In this paper, we will first measure systematically all kinds of automatic evaluation metrics over the same experimental setting to check which kind is best. 
    Through extensive experiments, the learning-based metrics are demonstrated that they are the most effective evaluation metrics for open-domain generative dialogue systems. 
    Moreover, we observe that nearly all learning-based metrics depend on the negative sampling mechanism, 
    which obtains an extremely imbalanced and low-quality dataset to train a score model. 
    In order to address this issue, we propose a novel and feasible learning-based metric that can significantly improve  the correlation with human judgments by using augmented \textbf{PO}sitive samples and valuable \textbf{NE}gative samples, called PONE.  
    Extensive experiments demonstrate that our proposed evaluation method significantly outperforms the state-of-the-art learning-based evaluation methods, 
    with an average correlation improvement of 13.18\%. 
    In addition, we have publicly released the codes of our proposed method and state-of-the-art baselines\footnote{Github address is \url{https://github.com/gmftbyGMFTBY/PONE}.}
\end{abstract}

\section{Introduction}
Open-domain generative dialogue systems (i.e. chatbots) have attracted considerable attention
\cite{Ritter2011DataDrivenRG,Yao2017TowardsIC,Zhou2018CommonsenseKA,Zhang2019ReCoSaDT}. 
Recent advances in open-domain generative dialogue systems highlight the difficulties in
automatic evaluation \cite{Kannan2016AdversarialEO,Tong2018OneF,Huang2019ChallengesIB}. 
Like other Natural Language Generation tasks, 
it is very difficult to evaluate open-domain generative dialogue systems 
because of the diversity of languages \cite{Gatt2017SurveyOT}.

Generally, human annotation is the most reliable evaluation method. 
However, the human annotation is very expensive, time-consuming and irreproducible. 
Thus, it is highly desirable to automatically evaluate the open-domain generative dialogue systems.  
To the best of our knowledge, so far, there are three kinds of metrics to automatically evaluate the open-domain generative dialogue systems: 
(1) Word-overlap-based metrics, they evaluate the amount
of word-overlap between a generated response and its corresponding ground-truth response, such as BLEU \cite{Papineni2001BleuAM} and ROUGE \cite{Lin2004ROUGEAP};
(2) Embedding-based metrics, they evaluate the quality of a generated response by 
calculating the semantic similarity between the generated response and its corresponding ground-truth,  
such as Greedy Matching \cite{Rus2012ACO} and BERTScore \cite{Zhang2019BERTScoreET};
(3) Learning-based metrics, they train a score model to evaluate a generated response based on its conversation context,  
such as ADEM \cite{Lowe2017TowardsAA} and RUBER \cite{Tao2018RUBERAU}; 
However, due to the lack of systematic comparison of these metrics, 
it is not clear which kind of automatic metrics are closer to human judgments, i.e., which kind is best.
In this paper, we will first measure systematically all kinds of automatic evaluation metrics  from three important aspects for open-domain generative dialogue systems: 
(1) Fluency: the grammatical correctness and fluency of generated responses; 
(2) Coherence: the relevence between the generated responses and their contexts.
(3) Engagement: the degree of attraction or engagement of generated responses \cite{ghazarian2019predictive}. 
Extensive systematic experiments demonstrate that the learning-based metrics are the most effective automatic evaluating metrics for open-domain dialogue systems,  and have a very high correlation with human judgments.  

Furthermore, we observe that nearly all learning-based metrics depend on the negative sampling mechanism to obtain datasets to train score models. 
Generally, in the obtained datasets:  
(1) The size of positive and negative samples is extremely imbalanced;
(2) Most negative samples are low-quality since they are randomly sampled.
Thus the decision boundary generated by these existing learning-based methods 
is far from the real decision boundary, which obviously hurts the performance.  
In order to narrow the difference between the real decision boundary and the generated decision boundary, 
we propose a novel learning-based evaluation method
by using augmented \textbf{PO}sitive samples and valuable \textbf{NE}gative samples to train the score model, called PONE.
Specifically, we first apply data augmentation techniques to obtain lots of generated positive samples.
Then, we adopt a novel sampling strategy to collect the valuable negative samples.
Finally, a novel iterative algorithm is proposed to overcome the impact of the noise in augmented data samples.
Extensive experiments demonstrate that our proposed method significantly outperforms the state-of-the-art learning-based metrics, 
with an average correlation improvement of 13.18\%.

In the following sections, we will first evaluate systematically all kinds of automatic metrics for open-domain dialogue systems. 
Then, we will introduce the detail of our proposed novel learning-based metric. 

\section{Systematic Comparison} \label{systematical}
In this section, we systematically compare the all kinds of automatic metrics. 

\begin{table}[h!]
    \small
    \begin{center}
        \resizebox{0.5\textwidth}{!}{
            \begin{tabular}{l|c|c|c|c}
            \toprule[2pt]
            \hline
            \textbf{Dataset} & \textbf{Fluency} & \textbf{Coherence} & \textbf{Engagement} & \textbf{Overall} \\ \hline
            Tencent          & 0.331 & 0.623 & 0.566 & 0.610 \\ \hline
            Xiaohuangji      & 0.292 & 0.632 & 0.625 & 0.627 \\ \hline
            Dailydialog      & 0.330 & 0.612 & 0.326 & 0.523 \\ \hline
            Cornell          & 0.249 & 0.468 & 0.388 & 0.343 \\ \hline
            \bottomrule[2pt]
            \end{tabular}
        }
        \caption{Inter-annotator Kappa agreement is high.}
        \label{tab:15}
    \end{center}
\end{table} 

\subsection{Experimental setup}
The overall evaluating process are as follows:
(1) 
To obtain the a response and its context, we need to choose the open-domain generative dialogue systems.
In order to make the conclusions more reliable, three generative models are chosen 
according to the difference of the decoding process:
(a) Seq2Seq-attn \cite{Bahdanau2014NeuralMT}; 
(b) Copynet \cite{Gu2016IncorporatingCM}; 
(c) Transformer \cite{Vaswani2017AttentionIA}. 
More details can be found in \textit{Appendix}. 
(2) 100 samples are randomly sampled from these generated pairs of the response and context.
Then the human judgements are obtained, $S_{human}$. Specifically, 3 volunteers are asked not only to provide an overall scores but also to provide scores from fluency, coherence, and engagement.
We ask the volunteers to rate each pair of the context and response on a scale of 1-6 (very bad to very good).
The inter-annotator Kappa value is show in Table \ref{tab:15}.
(3) All kinds of the automatic evaluation metrics are used to obtain the scores of the samples, $S_{metric}$.
(4) In order to measure the correlation between these automatic evaluation metrics and the human judgments,
Pearson and Spearman correlation are adopted to obtain the correlation between $S_{human}$ and $S_{metric}$.
Pearson and Spearman correlations estimate linear and monotonic correlation 
and are widely used in the automatic evaluation of open-domain generative dialogue systems 
such as RUBER \cite{Tao2018RUBERAU} and BERT-RUBER \cite{Ghazarian2019BetterAE}.
The higher the correlation between $S_{human}$ and $S_{metric}$, the closer the metric to human judgments.
If the $p$-value is less than 0.01, 
the relationship between human judgments and the automatic metrics is significant.

The details of the evaluating process are shown below.

\subsubsection{Chosen datasets}
\begin{itemize}
    \item \textbf{Tencent}: Tencent dataset \cite{Li2018AMA} contains high-quality open-domain conversation from social media.
    Each reply is annotated by humans according to the quality of the response.
    \item \textbf{Xiaohuangji\footnote{https://github.com/fate233/dgk\_lost\_conv}}: Xiaohuangji is a classic Chinese open-domain dialogue dataset with lots of colloquial interaction.  
    \item \textbf{Dailydialog}: Dailydialog dataset \cite{Li2017DailyDialogAM} is a multi-turn high-quality English open-domain dialogue corpus which covers various topics about our daily life.
    In this section, we process it into the single-turn format.
    \item \textbf{Cornell}: Cornell corpus \cite{Danescu-Niculescu-Mizil+Lee:11a} is an English chit-chat corpus which contains a large metadata-rich collection of fictional conversations extracted from raw movie scripts.
\end{itemize}

For open-domain generative dialogue models, we sample 90,000/2,500/2,500 query-response pairs for each corpus as train/test/validation sets.
For learning-based metrics, we sample 45,000/2,500/2,500 query-response pairs for each corpus as train/test/validation sets.


\begin{table*}[t!]
    \begin{center}
        \small
        \resizebox{\textwidth}{!}{
            \begin{tabular}{|c|c|c|c|c|c|c|c|c|c|}
            \toprule[2pt]
            \hline
            \multicolumn{2}{|c|}{\multirow{2}{*}{\textbf{Metrics}}}             & \multicolumn{2}{c|}{\textbf{Seq2Seq-attn Dailydialog}}                                                              & \multicolumn{2}{c|}{\textbf{Seq2Seq-atnn Cornell}}                                                            & \multicolumn{2}{c|}{\textbf{Transformer Dailydialog}}                                                           & \multicolumn{2}{c|}{\textbf{Transformer Cornell}}                                                              \\ \cline{3-10} 
            \multicolumn{2}{|c|}{}  & \multicolumn{1}{c|}{\textbf{Pearson (\textit{p})}} & \multicolumn{1}{c|}{\textbf{Spearman (\textit{p})}} & \multicolumn{1}{c|}{\textbf{Pearson (\textit{p})}} & \multicolumn{1}{c|}{\textbf{Spearman (\textit{p})}} & \multicolumn{1}{c|}{\textbf{Pearson (\textit{p})}} & \multicolumn{1}{c|}{\textbf{Spearman (\textit{p})}} & \multicolumn{1}{c|}{\textbf{Pearson (\textit{p})}} & \multicolumn{1}{c|}{\textbf{Spearman (\textit{p})}} \\ \hline \hline
            \multirow{2}{*}{\textbf{Human}}        & \textbf{Human(Avg)} & 0.3088 (0.0)    & 0.3078 (0.0)    & 0.4086 (0.0)    & 0.3903 (0.0)    & 0.42932(0.0004)	&0.4116(0.00117)    & 0.13636(0.29495)&	0.12885(0.37743) \\ \cline{2-10} 
                                                   & \textbf{Human(Max)} & 0.5979 (0.0)    & 0.2856 (0.0)    & 0.4285 (0.0)    & 0.4073 (0.0)    & 0.51501(0.00119)&	0.51607(0.0035)    & 0.2349(0.67459)&	0.2354(0.9706) \\ \hline \hline
            \multirow{6}{*}{\textbf{Word-overlap}} & \textbf{ROUGE}      & 0.2511 (0.0117) & 0.1901 (0.0581) & 0.0636 (0.5298) & 0.0423 (0.6759) & 0.34463(0.00045)&	0.36092(0.00023) & 0.22317(0.02562)&	0.246(0.01362) \\ \cline{2-10} 
                                                   & \textbf{BLEU-1}     & 0.0508 (0.2360) & 0.0407 (0.3269) & 0.2348 (0.0187) & 0.1854 (0.0648) & 0.12453(0.21703)&	0.14693(0.14463)& 0.0338(0.73852)&	0.13777(0.17164) \\ \cline{2-10} 
                                                   & \textbf{BLEU-2}     & -0.0536 (0.1829)& 0.0259 (0.1674) & 0.2099 (0.0361) & 0.1873 (0.0620) & 0.126(0.2116)&	0.13663(0.17526)& 0.12331(0.2216)	&0.12939(0.19949)\\ \cline{2-10} 
                                                   & \textbf{BLEU-3}     & -0.0958 (0.1264)& -0.0020 (0.1413)& 0.1943 (0.0528) & 0.1807 (0.0720) & 0.12113(0.22996)&	0.11568(0.25174)& 0.13633(0.17622)	&0.1386(0.16905)\\ \cline{2-10} 
                                                   & \textbf{BLEU-4}     & -0.0971 (0.1071)& -0.0227 (0.1900)& 0.1868 (0.0627) & 0.1704 (0.0902) & 0.11824(0.24133)&	0.09571(0.34353)& 0.13947(0.16639)&	0.15311(0.1283) \\ \cline{2-10} 
                                                   & \textbf{METEOR}     & 0.2350 (0.0186) & 0.1732 (0.0848) & 0.1558 (0.1217) & 0.2260 (0.0238) & 0.21651(0.0305) &0.08771(0.38553) & 0.15578(0.12171) &0.21586(0.03101) \\ \hline \hline
            \multirow{4}{*}{\textbf{Embedding}}    & \textbf{EA}         & 0.3487 (0.0004) & 0.3187 (0.0012) & 0.1222 (0.2259) & 0.1213 (0.2293) & 0.1983(0.04796)	&0.2573(0.00976) & 0.0409(0.68616)	&0.02308(0.8197) \\ \cline{2-10} 
                                                   & \textbf{VX}         & 0.3463 (0.0004) & 0.3744 (0.0001) & 0.0340 (0.7372) & 0.0310 (0.7594) & 0.15127(0.133)&	0.26948(0.0067) & -0.05111(0.61353)	&-0.07187(0.47733) \\ \cline{2-10} 
                                                   & \textbf{GM}         & 0.2548 (0.0105) & 0.2382 (0.0170) & 0.0507 (0.6167) & -0.0135 (0.8941)& 0.36146(0.00022)	&0.28486(0.00407) & -0.0349(0.73034)&	0.00193(0.98481) \\ \cline{2-10} 
                                                   & \textbf{BERTScore}  & 0.2049 (0.0408) & 0.2212 (0.0270) & 0.2648 (0.0078) & \textbf{0.3145} (0.0011) & 0.17809 (0.07628) &0.19813 (0.04815) & 0.22948 (0.02164) &0.26861 (0.00689) \\ \hline \hline
            \multirow{1}{*}{\textbf{Learning}} & \textbf{BERT-RUBER}     & \textbf{0.4000} (0.0002) & \textbf{0.4247} (0.0)    & \textbf{0.3193} (0.0034) & 0.3021 (0.0060) & \textbf{0.47541} (0.0)&	\textbf{0.51761(0.0)} & \textbf{0.35327} (0.00099)&	\textbf{0.343} (0.00129)\\ \hline \bottomrule[2pt]
            \end{tabular}
        }
        \caption{The correlation between automatic metrics and human judgments on Dailydialog dataset and Cornell dataset for Seq2Seq-attn and Transformer model. 
        \textit{p} means the \textit{p}-value. \textbf{EA}, \textbf{VX}, \textbf{GM} are the Embedding Average, Vector Extrema and Greedy Matching methods.
        We also compare human-human agreement: ``Human (Avg)" refers to average correlation between every two humans, whereas ``Human (Max)" refers to the two annotators who are most correlated. Best results are shown in bold.}
        \label{tab:1}
    \end{center}
\end{table*}

\begin{table*}[t!]
    \begin{center}
        \small
        \resizebox{\textwidth}{!}{
            \begin{tabular}{|c|c|c|c|c|c|c|c|}
            \toprule[2pt]
            \hline
            \multicolumn{2}{|c|}{\multirow{2}{*}{\textbf{Metrics}}}             & \multicolumn{2}{c|}{\textbf{Fluency}}                                                              & \multicolumn{2}{c|}{\textbf{Coherence}}                                                            & \multicolumn{2}{c|}{\textbf{Engagement}}                                                  \\ \cline{3-8} 
            \multicolumn{2}{|c|}{}  & \multicolumn{1}{c|}{\textbf{Pearson (\textit{p})}} & \multicolumn{1}{c|}{\textbf{Spearman (\textit{p})}} & \multicolumn{1}{c|}{\textbf{Pearson (\textit{p})}} & \multicolumn{1}{c|}{\textbf{Spearman (\textit{p})}} & \multicolumn{1}{c|}{\textbf{Pearson (\textit{p})}} & \multicolumn{1}{c|}{\textbf{Spearman (\textit{p})}} \\ \hline \hline
            \multirow{2}{*}{\textbf{Human}}        & \textbf{Human(Avg)} & 0.25671 (0.04433)&	0.23077 (0.04018)	&0.58819 (0.0)&	0.56235 (0.0)	&0.17629 (0.47127)&	0.17013 (0.48079)    \\ \cline{2-8} 
                                                   & \textbf{Human(Max)} & 0.35980 (0.12222)&	0.28559 (0.10125)&	0.71379 (0.0)	&0.68923 (0.0)&	0.44772 (0.96780)&	0.43289 (0.99546)    \\ \hline \hline
            \multirow{6}{*}{\textbf{Word-overlap}} & \textbf{ROUGE}      & 0.22758 (0.02277)	&0.16664 (0.09751)&	0.29557 (0.00283)&	0.20947 (0.03647)	&0.22673 (0.0233)&	0.11552 (0.25242) \\ \cline{2-8} 
                                                   & \textbf{BLEU-1}     & 0.16335 (0.10397)	& 0.10365 (0.30478)	&0.00471 (0.51825)	&0.02984 (0.65568)	&-0.0345 (0.07609)&	-0.02819 (0.0582)\\ \cline{2-8} 
                                                   & \textbf{BLEU-2}     & 0.14949 (0.15892)	&0.04786 (0.6363)&	-0.03278 (0.16866)&	0.03422 (0.63277)	&0.09611 (0.34147)&	0.02878 (0.16029)\\ \cline{2-8} 
                                                   & \textbf{BLEU-3}     & 0.12854 (0.20249)	&0.00941 (0.8566)&	-0.0502 (0.9956)&	0.01907 (0.52688)&	0.13893 (0.16805)&	0.07762 (0.31379)\\ \cline{2-8} 
                                                   & \textbf{BLEU-4}     & 0.12195 (0.22677)&	-0.0206 (0.6924)&	-0.05859 (0.07712)&	0.01682 (0.45016)	&0.1567 (0.11948)&	0.13136 (0.19268)\\ \cline{2-8} 
                                                   & \textbf{METEOR}     & 0.21733 (0.02986)&	0.15244 (0.13001)&	0.29886 (0.00252)&	0.18758 (0.06164)&	0.26144 (0.0086)&	0.17211 (0.08684) \\ \hline \hline
            \multirow{4}{*}{\textbf{Embedding}}    & \textbf{EA}         & 0.27328 (0.00594)	&0.25212 (0.01139)	&0.37456 (0.00012)	&0.34373 (0.00046)&	0.28161 (0.00453)&	0.25026 (0.01203) \\ \cline{2-8} 
                                                   & \textbf{VX}         & 0.16257 (0.01954)&	0.17511 (0.0814)&	0.41057 (2e-05)	&0.39348 (5e-5)&	0.20061 (0.04536)&	0.19938 (0.04673) \\ \cline{2-8} 
                                                   & \textbf{GM}         & 0.11472 (0.20558)&	0.06597 (0.27803)	&0.18876 (0.06)	&0.12979 (0.19808)&	0.19689 (0.04959)&	0.14733 (0.14354) \\ \cline{2-8} 
                                                   & \textbf{BERTScore}  & \textbf{0.49783} (0.0)&	0.19963 (0.04645)&	0.23521 (0.01849)&	0.33128 (0.00076)&	0.28702 (0.00379)&	0.16085 (0.10988) \\ \hline \hline
            \multirow{1}{*}{\textbf{Learning}} & \textbf{BERT-RUBER}     & 0.35603 (0.00075)&	\textbf{0.38392} (0.00019)&	\textbf{0.37732} (0.00086)&	\textbf{0.41318} (0.00011)&	\textbf{0.35303} (0.00133)	&\textbf{0.38185} (0.00029) \\ \hline \bottomrule[2pt]
            \end{tabular}
        }
        \caption{The fluency, coherence, engagement correlation between automatic metrics and human judgments on Dailydialog dataset for the Seq2Seq-attn model. Best results are shown in bold.}
        \label{tab:2}
    \end{center}
\end{table*}


\subsubsection{Chosen automatic evaluation metrics}
Three kind of metrics will be chosen to evaluate:
\begin{itemize}
    \item Word-overlap-based: BLEU \cite{Papineni2001BleuAM}, ROUGE \cite{Lin2004ROUGEAP} and METEOR \cite{Banerjee2005METEORAA} are the most commonly used automatic evaluation for open-domain generative dialogue systems.
    \item Embedding-based: Four embedding-based metrics are selected: (1) Embedding-Average \cite{Wieting2015TowardsUP}; (2) Vector-Extrema \cite{Forgues2014BootstrappingDS}; (3) Greedy-Matching\footnote{We adopt Glove \cite{Pennington2014GloveGV} as English word embeddings. Chinese word embeddings are from https://github.com/Embedding/Chinese-Word-Vectors.} \cite{Rus2012ACO}; (4) BERTScore \cite{Zhang2019BERTScoreET}. 
    It should be noted that the BERTScore is the state-of-the-art embedding-based evaluation method which is based on BERT contextual embeddings\footnote{We use lastest version 0.2 of BERTScore.}.
    \item Learning-based: 
    So far, the feasible learning-based metrics are RUBER \cite{Tao2018RUBERAU} and BERT-RUBER \cite{Ghazarian2019BetterAE}. 
    RUBER is an unfinished approach, and its performance is very bad in most cases.
    BERT-RUBER overcomes the drawback of the RUBER, is a representative learning-based metric.
    Thus, BERT-RUBER is chosen here.
\end{itemize}



\subsection{Results}
Extensive experiments are conducted for all kinds of metrics which leads to 12 tables. 
Due to the page limitation, we only show the part of them. 
More details can be found in \textit{Appendix}.
Noted that the conclusions over these partial tables here are consistent with all tables.
We can make the following observations:
\begin{itemize}
    \item As shown in Table \ref{tab:1}, it can be found that the learning-based metric is much better than the embedding-based and word-overlap-based metrics in most cases, 
    and the learning-based metric shows a very close correlation with human judgments. 
    \item We also test these three kinds of automatic evaluation metrics based on the three important aspects of evaluating dialogue systems: (1) Fluency; (2) Coherence; (3) Engagement.
    As shown in Table \ref{tab:2}, it can be found that the learning-based metric achieves the closest correlation with human judgments on these aspects in most cases. 
    Because learning-based metrics are highly relevant to human judgments on these important aspects of evaluating the dialogue systems, 
    the scores predicted by learning-based metric are very reliable.
\end{itemize}

Learning-based metrics are very powerful to evaluate the generative dialogue systems.
\textbf{But, are the learning-based metrics really good enough?}

\section{Proposed metric} \label{propose}
In this section, we first analyze the fatal weakness of the existing learning-based metrics,
and then propose a novel learning-based automatic evaluation of open-domain dialogue systems which alleviates the issue.
Finally, we show the comparison between the existing learning-based metrics 
on single-turn and multi-turn dialogue benchmarks.
It should be noted that unlike the traditional NLP tasks, 
automatic evaluations need very high practicability, 
so the automatic evaluation approaches are simple but very effective (complex models are hard to obtain).

\begin{figure}[t]
    \centering
    \subfigure[Traditional negative sampling.]{
        \begin{minipage}[t]{0.5\linewidth}
            \centering
            \includegraphics[width=4cm, height=4cm]{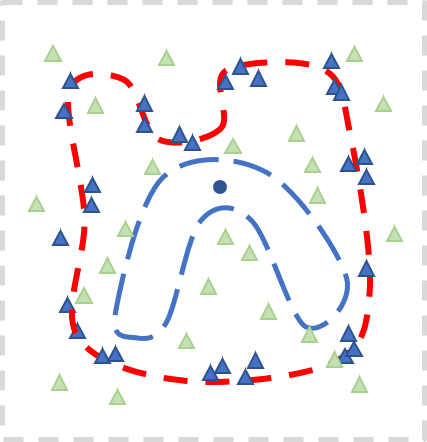}
        \end{minipage}%
        \label{img:1a}
    }%
    \subfigure[PONE.]{
        \begin{minipage}[t]{0.5\linewidth}
            \centering
            \includegraphics[width=4cm, height=4cm]{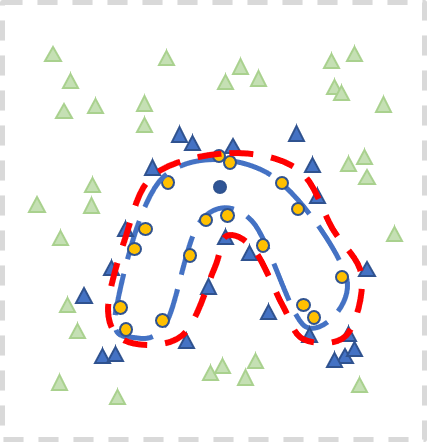}
        \end{minipage}%
        \label{img:1b}
    }%
    \centering
    \caption{\small The examples of the decision boundary and data points. Dots are positive sample and triangles are negative samples. The real decision boundary is dashed blue line and the generated one is red.
    In Figure 1 (a), only one positve sample and random negative samples generate the not so good decision boundary. In Figure 1 (b), the decision boundary generated by our proposed method is closer to the real decision boundary.}
    \label{img:1}
\end{figure}

\subsection{Weakness and motivation}
The learning-based metrics train the score model by using negative sampling.
Specifically, the ground-truth of the a query is the positive sample, and the randomly sampled responses are the negative samples.
These samples are used to train a binary classifier (score model), which will be used to predict the score by \textit{sigmoid} function.
However, the obtained dataset generated by the learning-based metrics has two drawbacks: 
(1) The size of positive and negative samples is extremely imbalanced. 
As shown in Figure \ref{img:1} (a), given a context, there is only one positive response and 
lots of negative samples;
(2) Most negative samples are low-quality since they are randomly sampled. 
So the decision boundary generated by these existing learning-based methods 
is far from the real decision boundary, which hurts the performance.

Intuitively, as shown in Figure \ref{img:1} (b), the real decision boundary can be highlighted by lots of positive samples and 
the valuable negative samples which are close to positive samples. 
In order to narrow the difference between the real decision boundary and the generated decision boundary, 
we propose a novel evaluation method
by using augmented \textbf{PO}sitive samples and valuable \textbf{NE}gative samples to train the score model, called PONE.
Specifically, we first generate lots of effective positive samples.
Then, we apply a novel sampling strategy based on BERT contextual embedding to collect the valuable negative samples for training. 
Besides, we also propose a novel iterative algorithm to decrease the impact of the noise in generated positive samples.

\subsection{Notations and task formulation}
Given a response $r$ and the corresponding context query $q$,
an evaluating metric needs to predict the score $s$ of $r$ based on the $q$. 
The score $s$ ranges from 0 to 1 (including 0 and 1), 
where higher scores indicate higher quality of the generated responses. 
In this task, we have to make sure that the output of the proposed metric 
is highly relevant to human judgments.

\subsection{The architecture of PONE}
Our proposed evaluation method PONE contains four components:
(1) Weighted negative sampler, it selects the valuable negative samples instead of random samples;
(2) Positive data generator, it provides lots of effective positive samples 
which alleviates the imbalanced data distribution;
(3) Label filter, it is a novel algorithm for detecting noise augmented data;
(4) Scorer, it is score model (classifier) of the PONE for predicting the score which indicates the quality of the responses. 

\subsubsection{Weighted negative sampler}

\begin{table}[h]
    \begin{center}
        \resizebox{0.5\textwidth}{!}{
            \begin{tabular}{ccc}
            \toprule[2pt]
            \hline
            \makecell[c]{\textbf{Ground-truth reply}} & \makecell[c]{\textbf{Weighted negative} \\ \textbf{samples}} & \makecell[c]{\textbf{Random negative} \\ \textbf{samples}} \\ \hline
            \makecell[c]{My wife is coming\\ to get me.} & \makecell[c]{Yes. She is meeting\\ a client.} & \makecell[c]{Are you serious?}       \\ \hline
            \makecell[c]{Weather reports say it is \\ going to rain for a week.} & \makecell[c]{It may rain again\\ later today.} & \makecell[c]{It has been conquered \\ by many people \\ in history.}       \\ \hline
            \makecell[c]{I'm an engineer at IBM.} & \makecell[c]{I used to \\ work as a teacher.} & \makecell[c]{Wow. \\ You are really \\ a heavy smoker .}       \\ \hline
            \bottomrule[2pt]
            \end{tabular}
        }
        \caption{Real examples from Dailydialog dataset, weighted negative samples are better than random negative samples.}
        \label{tab:3}
    \end{center}
\end{table}

As shown in Table \ref{tab:3}, most of the random negative samples can be easily discriminated from positive samples and will contribute little towards the training \cite{Cai2017KBGANAL}. High-quality negative samples that are close to the ground-truth are beneficial to approximate the real decision boundary.
So it is necessary to select valuable negative samples for training.
First, we randomly extract $h$ candidate negative samples $\{n_i\}_{i=1}^{h}$ from the whole datasets.
This operation is necessary because we have to speed up the training process. 
Then we obtain the similarities between the ground-truth $r$ and the candidate negative samples $\{n_i\}_{i=1}^h$. 
Finally, we convert the semantic similarity to the sample probability which 
will be used to select the closest candidate sentence $n_i$ to the ground-truth $r$ as the negative sample.

During training, we adopt the average pooling strategy to get the BERT \cite{devlin2018bert} 
semantic embeddings $\bm{v}_r$, $\bm{v}_{n_i}$, $\bm{v}_q$ of the ground-truth response $r$, negative response $n_i$ and context $q$ respectively.
Formally, let $w_1$, $w_2$, ..., $w_m$ be the BERT embedding of $m$ words in an utterance.
Average pooling takes the average of the hidden state of the encoding layer on the time axis.
\begin{equation} \label{form:1}
    \begin{split}
        & \bm{v}=\rm{avg} \{w_1[i], w_2[i], ..., w_m[i]\}\\
    \end{split}
\end{equation} where $[\cdot]$ indexes a dimension of a vector\footnote{We use the last layer of the BERT architecture, and the embedding size is 768. Concatenation of the last four layers turns to the bad performance.}. Then we obtain the semantic similarity of the ground-truth $r$ and negative sample $n_i$ by adopting cosine similarity:
\begin{equation} \label{form:2}
    \begin{split}
        & cos(\bm{v}_r,\bm{v}_{n_i})=\frac{\bm{v}_r^T\bm{v}_{n_i}}{\|\bm{v}_r\|\|\bm{v}_{n_i}\|}\\
    \end{split}
\end{equation} After getting all the cosine similarities of the candidate negative samples, 
we convert the cosine similarity to the selective probability by 
applying the \textit{softmax} with temperature factor.
\begin{equation} \label{form:3}
    \begin{split}
        & p(\bm{v}_r,\bm{v}_{n_i})=\frac{e^{cos(\bm{v}_r, \bm{v}_{n_i})/t}}{\sum_i^h e^{cos(\bm{v}_r, \bm{v}_{n_i})/t}}\\
    \end{split}
\end{equation} where $t$ is the temperature factor, $n_i$ is the $i$-th candidate negative sample.
$p(\bm{v}_r, \bm{v}_{n_i})$ indicates the probability that the $i$-th candidate negative sample is sampled.
We study the impact of the $t$ in the experiment section.

\subsubsection{Positive data generator}
All of the existing learning-based evaluation methods apply the negative sampling method. But the negative sampling algorithm constructs the extremely imbalanced dataset for training the score model. 
In order to address this issue, we apply the data augmentation approaches to generate lots of effective positive samples.
We test two kinds of data augmentation methods: (1) replacement-based method, i.e., EDA \cite{Wei2019EDAED}, it can generate extensive samples very fast by using four simple but powerful operations (synonym replacement, random insertion, random swap and random deletion). 
The main drawback of the EDA data augmentation is that the generated samples will contain lots of noise, which may hurt the performance; 
(2) generation-based method, i.e., Seq2Seq \cite{Cho2014LearningPR}, generative dialogue model Seq2Seq is used to generate the samples by using beam search. 
The generated samples are high-quality but it runs slowly.

After the data augmentation, the diverse augmented positive samples $\{r_{ij}\}_{j=1}^k$ based on context $q_i$ can be obtained, and its size of it is $k$.

\subsubsection{Scorer} 


The scorer is the score model of the proposed learning-based metric.
First, we obtain the semantic representation of the context $\bm{v}_q$ and response $\bm{v}_r$ from the dataset generated by the positive data generator and weighted negative sampler.
In this paper, we choose to apply BERT contextual embeddings to obtain the semantic representation of the sentences.
Then we further concatenate $\bm{v}_q$ and $\bm{v}_r$ to match the two utterances. 
Besides, we also include the bilinear projector to hold interactive information of $\bm{v}_q$ and $\bm{v}_{r}$ 
which is the same as the RUBER and BERT-RUBER.
Finally, a multi-layer perceptron (MLP) predicts a score as the final metric $s$. 
The hidden layers of MLP use $\textit{relu}$ as the activation function, whereas the last unit uses
$\textit{sigmoid}$ because we hope the score is a real number which is between 0 and 1.
The training objective is to minimize binary cross-entropy loss:
\begin{equation} \label{4}
    \begin{split}
        & J = \frac{1}{K} \sum_i^{K} -y_i \cdot \log \overline{y_i} - (1-y_i) \cdot \log (1-\overline{y_i}) \\
    \end{split}
\end{equation} where $K$ is the size of the whole generated dataset, $y_i$ is the label of the $i$-th example, 
$\overline{y_i}$ is the label of $i$-th example generated by the scorer.

\subsubsection{Label filter}
Actually, the augmented positive samples will contain lots of the noise inevitably \cite{Wei2019EDAED}.
In order to reduce the effect of noise samples to the performance of the scorer, 
we need to assign the real label for each augmented sample as many as possible. 
Fortunately, this task is essentially the same as the evaluation of open-domain dialogue systems. 
So we propose a novel iterative algorithm denoted as the label filter to reduce the impact of the noise samples.

The inputs to the algorithm are positive samples (ground-truth and augmented samples) and weighted negative samples. 
First, we use the positive and negative samples to pre-train the scorer, the formula is shown in \ref{4}. 
Then we apply the pre-trained scorer to generate pseudo labels on positive samples generated by the positive data generator. 
Furthermore, the scorer is fine-tuned by the negative samples and the augmented positive samples (with pseudo labels).
Finally, we iterate the process until the pseudo labels of the augmented samples no longer change.
The details of the algorithm can be found in the \textit{Appendix}.


During the training iteration, the scorer can generate more high-quality pseudo labels for each augmented positive samples which is beneficial to improve the performance further. 

\subsection{Experiments}
We compare PONE with existing state-of-the-art learning-based metrics by the correlation with human judgments.
In order to make the conclusions more general and convincing, 
we also extend the experiments to multi-turn dialogue models.

\subsubsection{Datasets}
For the single-turn setting, the experiment settings are the same as the Section \ref{systematical}.
For the multi-turn setting, the Dailydialog benchmark is used.
It should be noted that the Dailydialog is a multi-turn open-domain dataset and 
we process it into the single-turn and multi-turn corpus separately.
It should be noted that another 90,000/2,500/2,500 pairs for each corpus are used to train the data augmentation model Seq2Seq.

\subsubsection{Baselines}
We compare the proposed metric PONE with the state-of-the-art learning-based metric BERT-RUBER. 
BERT-RUBER trains the score model by negative sampling which achieves the state-of-the-art performance.
We also verify the effectiveness of PONE's components.
\begin{itemize}
    \item PONE-Po-LF: Without the positive data generator and label filter, only the weighted negative sampler is used to select valuable negative samples.
    \item PONE-Ne-LF: Only the positive data generator is applied to generate lots of meaningful positive samples. Seq2Seq data augmentation method is used by default.
    \item PONE-Ne: Remove the weighted negative sampler from the PONE.
\end{itemize}


\subsubsection{Generative dialogue models}
For single-turn dialogue systems, the settings are the same as the Section \ref{systematical}.
As for multi-turn dialogue systems, in order to make the conclusions more reliable, we apply two novel multi-turn dialogue models to generate the responses: 
(1) HRED \cite{Serban2015BuildingED}, HRED is the first hierarchical encoder-decoder architecture, and has been widely applied for multi-turn dialogue generation;
(2) ReCoSa \cite{Zhang2019ReCoSaDT}, ReCoSa adopts the self-attention mechanism in multi-turn dialogue modeling and achieve the state-of-the-art performance.
More details can be found in \textit{Appendix}.
The setup of single-turn dialogue systems is the same as the Section \ref{systematical}.

\subsubsection{Experiment Setup}
Parameters and details of our proposed evaluation method are shown in Table \ref{tab:4}.
Other details of the experiments are the same as the Section \ref{systematical}. 
300 randomly sampled utterances are annotated by 3 annotators for multi-turn dialogue systems.

\begin{table}[h!]
    \small
    \begin{center}
        \resizebox{0.5\textwidth}{!}{
            \begin{tabular}{|l|l|l|l|}
            \toprule[2pt]
            \hline
            \textbf{Param} & \textbf{Value} & \textbf{Param} & \textbf{Value} \\ \hline
            BERT embedding & 768            & $k$            & 5              \\ \hline
            BERT Pooling   & Mean           & epochs         & 100            \\ \hline
            Dropout ratio  & 0.5            & learning rate  & 1e-3           \\ \hline
            MLP $h$        & 256/512/128/1  & Optimizer      & Adam           \\ \hline 
            \bottomrule[2pt]
            \end{tabular}
        }
        \caption{Parameters setting of the PONE, $h$ represents the hidden size.}
        \label{tab:4}
    \end{center}
\end{table}

\begin{table*}[t!]
    \begin{center}
        \resizebox{\textwidth}{!}{
            \begin{tabular}{llllllll}
            \toprule[2pt]
            \hline
            \multicolumn{1}{c}{\textbf{Query}} & \multicolumn{1}{c}{\textbf{Ground-truth}} & \multicolumn{1}{c}{\textbf{Generated Response}} & \multicolumn{1}{c}{\textbf{BLEU}} & \multicolumn{1}{c}{\textbf{BERTScore}} & \multicolumn{1}{c}{\textbf{BERT-RUBER}} & \multicolumn{1}{c}{\textbf{PONE}} & \multicolumn{1}{c}{\textbf{Human}} \\ \hline
            
            \makecell[c]{Tell me a little bit about \\yourself please.} & 
            \makecell[c]{I'm Dunlin and I live in Beijing.\\ I was born in 1980.} & 
            \makecell[c]{Ok, thank you very much} & 
            \makecell[c]{0.1988} & 
            \makecell[c]{0.8168} & 
            \makecell[c]{0.9908} & 
            \makecell[c]{0.2903} & 
            \makecell[c]{0.20} \\ \hline
            
            \makecell[c]{I can type 100 Chinese words\\  a minute and take dictation \\ in English at 150 words a minute.} & 
            \makecell[c]{Can you operate \\computers skillfully?} & 
            \makecell[c]{So you don't have a printer?} & 
            \makecell[c]{0.2325} & 
            \makecell[c]{0.8793} & 
            \makecell[c]{0.5325} & 
            \makecell[c]{0.0554} & 
            \makecell[c]{0.0667} \\ \hline
            
            \makecell[c]{I usually have my lunch here. \\The food is tasty and \\ they offer a big variety.} & 
            \makecell[c]{Let's line up. \\The steak is very inviting.} & 
            \makecell[c]{That sounds pretty good.\\ Can you come here \\ for a few minutes?} & 
            \makecell[c]{0.2555} & 
            \makecell[c]{0.8566} & 
            \makecell[c]{0.7415} & 
            \makecell[c]{0.5631} & 
            \makecell[c]{0.5333} \\ \hline
            
            \makecell[c]{That sounds nice.} & 
            \makecell[c]{Yeah, you should go and\\ get it for yourself.} & 
            \makecell[c]{Yes, it is. \\ Let's go and have a drink .} & 
            \makecell[c]{0.1926} & 
            \makecell[c]{0.8861} & 
            \makecell[c]{0.8188} & 
            \makecell[c]{0.6814} & 
            \makecell[c]{0.60} \\
            
            \hline \bottomrule[2pt]                                   
            \end{tabular}
        }
        \caption{Real examples generated by Seq2Seq-attn model from Dailydialog dataset (single-turn). Human scores are the average of all the annotations and scores are normalized between 0 and 1.}
        \label{tab:6}
    \end{center}
\end{table*}


%

\begin{table*}[h]
    \begin{center}
        \resizebox{\textwidth}{!}{
            \begin{tabular}{|c|c|c|c|c|c|c|c|c|}
            \toprule[2pt]
            \hline
            \multirow{2}{*}{\textbf{Metrics}} & \multicolumn{2}{c|}{\textbf{CopyNet Dailydialog}}    & \multicolumn{2}{c|}{\textbf{CopyNet Xiaohuangji}}    & \multicolumn{2}{c|}{\textbf{Seq2Seq-attn Tencent}}  & \multicolumn{2}{c|}{\textbf{Seq2Seq-attn Dailydialog}}     \\ \cline{2-9} 
                                              & \textbf{Pearson (p)} & \textbf{Spearman (p)} & \textbf{Pearson (p)} & \textbf{Spearman (p)} & \textbf{Pearson (p)} & \textbf{Spearman (p)} & \textbf{Pearson (p)} & \textbf{Spearman (p)} \\ \hline
            \textbf{Human (max)}              &  0.65121(0.01416)    &  0.63276(0.02283)      &  0.65553(0.0)         &  0.63705(0.0)           & 0.74559 (0.0)          & 0.76104 (0.0)              & 0.59791 (0.0)   & 0.59002(0.0)      \\ \hline
            \textbf{BERTScore}                &  -0.1086(0.28216)    &  0.10627(0.29267)      &  0.11698(0.24642)     &  0.30987(0.0017)        & 0.40112(4e-5)          & 0.32937(0.00082)           & 0.20491 (0.04083) & 0.22119 (0.027)   \\ \hline \hline
            \textbf{BERT-RUBER}               &  0.36499(0.00049)    &  0.36142(0.00065)      &  0.28621(0.0096)      &  0.27972(0.0144)        & 0.42047 (1e-5)         & 0.41614 (2e-5)             & 0.40004 (0.00021) & 0.42647 (0.6e-5)     \\ \hline \hline
            \textbf{PONE-Po-LF}               &  0.41288(5e-05)      &  0.4122(4e-05)         &  0.38903(0.00044)     &  0.40812(0.0002)        & 0.47756 (0.0)          & 0.4523 (0.0)               & 0.41684 (4e-5) & 0.42858 (3e-5)   \\ \hline
            \textbf{PONE-Ne-LF}             &  0.43653(1e-5)       &  0.41926(1e-5)         &  0.45529(0.0)         &  0.49689 (0.0)          & 0.51411 (0.0)          & 0.49312 (0.0)              & 0.50408 (0.0)     & \textbf{0.51315} (0.0)               \\ \hline
            \textbf{PONE-Ne}              &  0.44077(1e-05)      &  0.41944(2e-05)        &  0.34813(0.00039)     &  0.34084(0.00052)       & \textbf{0.58183} (0.0) & 0.52846 (0.0)              & 0.51087 (0.0)    & 0.50825 (0.0)   \\ \hline
            \textbf{PONE}                     &  \textbf{0.47069}(0.0) &\textbf{0.46299}(0.0) &  \textbf{0.45649}(0.0)&  \textbf{0.4997}(0.0)   & 0.57168 (0.0) & \textbf{0.53006} (0.0)              & \textbf{0.51261} (0.0) & 0.50521 (0.0)   \\ \hline \bottomrule[2pt]
            \end{tabular}
        }
        \caption{The correlation between automatic metrics and human judgments for CopyNet and Transformer model. \textit{p} means the \textit{p}-value.
        Due to the unstable performance, RUBER cannot show the good performance. Data augmentation is Seq2Seq.}
        \label{tab:5}
    \end{center}
\end{table*}

\begin{table}[h]
    \begin{center}
        \resizebox{0.5\textwidth}{!}{
            \begin{tabular}{|c|c|c|c|c|}
            \toprule[2pt]
            \hline
            \multirow{2}{*}{\textbf{Metrics}} & \multicolumn{2}{c|}{\textbf{HRED}}    & \multicolumn{2}{c|}{\textbf{ReCoSa}}     \\ \cline{2-5} 
                                              & \textbf{Pearson (p)}  & \textbf{Spearman (p)} & \textbf{Pearson (p)}  & \textbf{Spearman (p)} \\ \hline
            \textbf{Human (max)}              &  0.6139 (0.0)         &  0.6924 (0.0)         &  0.6835 (0.0)         &  0.6830 (0.0)    \\ \hline
            \textbf{BERTScore}                &  0.3399 (0.0)         &  0.2880 (0.0)         &  0.2072 (0.0)         &  0.2178 (0.0)    \\ \hline \hline
            \textbf{BERT-RUBER}               &  0.3228 (0.0)         &  0.3026 (0.0)         &  0.24237 (5e-5)       &  0.25873 (1e-5)   \\ \hline \hline
            \textbf{PONE-Po-LF}               &  0.3432 (0.0)         &  0.3174 (0.0)         &  0.2850 (0.0)         &  0.2994 (0.0)    \\ \hline
            \textbf{PONE-Ne-LF}               &  0.3542 (0.0)         &  0.3233 (0.0)         &  0.29216 (0.0)        &  0.30537 (0.01754) \\ \hline
            \textbf{PONE-Ne}                  &  0.3741 (0.0)         &  0.3411 (0.0)        &  0.2867 (0.0)          &  0.30755 (0.0)   \\ \hline
            \textbf{PONE}                     &  \textbf{0.3743} (0.0)&  \textbf{0.3433} (0.0)&  \textbf{0.3018} (0.0)        &  \textbf{0.32066} (0.0) \\ \hline \bottomrule[2pt]
            \end{tabular}
        }
        \caption{Test the performance of the metrics on multi-turn open-domain dialogue systems. The correlation between automatic metrics and human judgments on Dailydialog dataset for HRED and ReCoSa model. \textit{p} means the \textit{p}-value.
        Data augmentation is the Seq2Seq.}
        \label{tab:7}
    \end{center}
\end{table}

\begin{table}[h]
    \begin{center}
        \resizebox{0.5\textwidth}{!}{
            \begin{tabular}{|c|c|c|c|c|}
            \toprule[2pt]
            \hline
            \multirow{2}{*}{\textbf{Metrics}} & \multicolumn{2}{c|}{\textbf{CopyNet Dailydialog}}    & \multicolumn{2}{c|}{\textbf{Seq2Seq-attn Dailydialog}}     \\ \cline{2-5} 
                                              & \textbf{Pearson (p)}  & \textbf{Spearman (p)} & \textbf{Pearson (p)}  & \textbf{Spearman (p)} \\ \hline
            \textbf{Human (max)}              &  0.6139 (0.0)         &  0.6924 (0.0)         &  0.6835 (0.0)         &  0.6830 (0.0)    \\ \hline
            \textbf{BERTScore}                &  0.3399 (0.0)         &  0.2880 (0.0)         &  0.2072 (0.0)         &  0.2178 (0.0)    \\ \hline \hline
            \textbf{BERT-RUBER}               &  0.36499(0.00049)      &  0.36142(0.00065)       &  0.40004 (0.00021) & 0.42647 (0.6e-5)   \\ \hline \hline
            \textbf{$\rm{PONE_{EDA}}$}                      &  \textbf{0.47452}(5e-05)        &  0.45376(4e-05)          &  0.46771 (0.3e-5)  & 0.47485 (1e-5)    \\ \hline
            \textbf{$\rm{PONE_{Seq2Seq}}$}                  &  0.47069(1e-5)&  \textbf{0.46299}(1e-5)    &  \textbf{0.51261} (0.0)     & \textbf{0.50521} (0.0)  \\ \hline
           \bottomrule[2pt]
           \end{tabular}
       }
       \caption{Different data augmentation technique of positive data augmentation. $\rm{PONE_{EDA}}$ means the EDA data augmentation technique is used in PONE.
       $\rm{PONE_{Seq2Seq}}$ means the Seq2Seq data augmentation is used in PONE.}
       \label{tab:8}
   \end{center}
\end{table}

\subsubsection{Results}
Due to the page limitation, we only show partial results of the experiment. 
More details can be found in the \textit{Appendix}.
The conclusions we observed in these tables are consistent.

\noindent \textbf{Single-turn:} As shown in Table \ref{tab:5}, it can be found that
the proposed metric PONE is much better than the state-of-the-art metric BERT-RUBER, 
with an average correlation improvement of 13.18\%. 
Besides, the performance of PONE is very close to the average human level.


\noindent \textbf{Multi-turn:} As shown in Table \ref{tab:7}, we can make the following conclusions:
(1) Our proposed metric PONE significantly outperforms the state-of-the-art metrics BERT-RUBER and BERTScore. 
(2) Compared with the performance on the single-turn setting, the performance of PONE is still far from the human average. 
We attribute this problem to the fuzzy semantic caused by the long context of multi-turn conversations context, 
which means that there is still an urgent need to explore the better metric for multi-turn dialogue systems.

\noindent \textbf{Ablation Study:} We also verify the effectiveness of three components in PONE's framework (last 4 rows in Table \ref{tab:5} and Table \ref{tab:7}). We can make the conclusions:
\begin{itemize}
    \item \textbf{Weighted negative sampler}: As shown in Table \ref{tab:5}, 
    the results of PONE-Po-LF demonstrate that 
    the weighted negative sampler really helps to improve the correlation with human judgments, compared with the state-of-the-art baseline BERT-RUBER. 
    This is because the weighted negative sampling approach can find appropriate negative samples to ensure that the binary classifier can learn valuable features that distinguish positive samples with negative samples.
    \item \textbf{Positive data generator}: 
    As shown in Table \ref{tab:5}, 
    it can be found that adopting the data augmentation methods improve the performance.
    We also compare the two data augmentation approaches. As shown in Table \ref{tab:8},
    $\rm{PONE_{EDA}}$ and $\rm{PONE_{Seq2Seq}}$ can both improve the performance, which demonstrates the data augmentation effectively alleviates the imbalanced dataset issue.
    Furthermore, it can be shown that applying the Seq2Seq data augmentation can achieve better performance than EDA.
    It demonstrates that the higher the quality of the generated samples, the greater the effect.
    \item \textbf{Label filter algorithm}: 
    Adopting the label filter algorithm for the positive data generator can further improve the performance in most cases. 
    It effectively alleviates the issue of the noise in the augmented positive samples.
    \item \textbf{PONE}: 
    As shown in Table \ref{tab:5}, it can be found that directly combining weighted negative sampler and positive data generator in PONE
    sometimes decreases the performance.
    It means that the way of combining two components needs to be further studied, and we will leave this part for future research.
\end{itemize}

\noindent \textbf{Hyperparameters}: We analyze two main hyperparameters in our work: 
(1) Temperature factor $t$, the performance is best when the $t$ is between 0.05 and 0.1; 
(2) The number of the generated positive samples $k$, the performance is best when $k$ is 5. 
More details can be found in \textit{Appendix}.

\subsubsection{Case Study}

Some real examples generated over the Dailydialog dataset are shown in Table \ref{tab:6}. 
We can make the conclusions as follows: 
(1) Because of occasional exact word overlapping, word-overlap-based metrics provide small scores, 
and it is not appropriate for evaluating the open-domain dialogue systems. 
We obtain the average score of the BLEU on all the datasets, and the result is 0.173; 
(2) The scores generated by the embedding-based metrics such as BERTScore and Embedding Average are usually very high 
because the sentence representation is very fuzzy. 
We also get the average scores of the embedding-based metrics on all datasets, and it is 0.814; 
(3) The learning-based metrics such as BERT-RUBER and PONE can provide balanced (average 0.473 of PONE) 
and, appropriate score and the PONE's scores are closer to the human judgments.

To test whether PONE generates a better decision boundary than BERT-RUBER, 
we also visualize the decision boundary on Dailydialog dataset.
Due to the page limitation, the visualized result can be found in \textit{Appendix}. 

\section{Conclusion}
In this paper, we conduct a comprehensive analysis of the existing evaluation methods of open-domain dialogue systems from fluency, coherence, engagement. 
Extensive experiments demonstrate that learning-based evaluation methods are much better than traditional metrics. 
Furthermore, we propose a novel learning-based automatic evaluation method, called PONE. 
The proposed metric significantly outperforms the state-of-the-art learning-based evaluation method. 
 

\bibliography{anthology,acl2020}
\bibliographystyle{acl_natbib}

\end{document}